\documentclass{article}
\usepackage[japanese]{babel}
\usepackage[T1]{fontenc} 
\usepackage[utf8]{inputenc} 
\usepackage{ismir,amsmath,cite}

\usepackage{graphicx}

\usepackage{color}
\usepackage{microtype}
\usepackage{xurl}
\usepackage{CJKutf8}

\usepackage{booktabs}
\usepackage{multirow}
\usepackage[table,xcdraw]{xcolor}
\usepackage[normalem]{ulem}
\useunder{\uline}{\ul}{}

\usepackage{lineno}

\newcommand{\concat}{%
  \mathbin{{+}\mspace{-8mu}{+}}%
}

\title{A Computational Evaluation Framework\\for Singable Lyric Translation}

\multauthor
{Haven Kim $^1$ \hspace{1cm} Kento Watanabe $^2$
\hspace{1cm}Masataka Goto $^2$ \hspace{1cm} Juhan Nam $^1$}
{
  $^1$ Graduate School of Culture Technology, KAIST, South Korea\\
$^2$ National Institute of Advanced Industrial Science and Technology (AIST), Japan\\
{\tt\small khaven@kaist.ac.kr, kento.watanabe@aist.go.jp, m.goto@aist.go.jp, juhan.nam@kaist.ac.kr}
}

\def\authorname{H. Kim, K. Watanabe, M. Goto, and J. Nam}

\begin{document}

\maketitle
\begin{abstract}
Lyric translation plays a pivotal role in amplifying the global resonance of music, bridging cultural divides, and fostering universal connections. Translating lyrics, unlike conventional translation tasks, requires a delicate balance between singability and semantics. In this paper, we present a computational framework for the quantitative evaluation of singable lyric translation, which seamlessly integrates musical, linguistic, and cultural dimensions of lyrics. Our comprehensive framework consists of four metrics that measure syllable count distance, phoneme repetition similarity, musical structure distance, and semantic similarity. To substantiate the efficacy of our framework, we collected a singable lyrics dataset, which precisely aligns English, Japanese, and Korean lyrics on a line-by-line and section-by-section basis, and conducted a comparative analysis between singable and non-singable lyrics. Our multidisciplinary approach provides insights into the key components that underlie the art of lyric translation and establishes a solid groundwork for the future of computational lyric translation assessment.


\end{abstract}
\begin{CJK}{UTF8}{mj}
\section{Introduction}\label{sec:introduction}
Translating lyrics is a prevailing method of enhancing the global appeal and allure of music across a multitude of genres, such as theatre music, animation music, pop music, etc~\cite{mateo2012music}. Furthermore, recent advancements in media technology have facilitated the exchange of intercultural products and globalized fandom culture, resulting in an increase in the popularity of user-translated lyrics across diverse social media platforms~\cite{susam2008translation}. 

Despite its popularity, lyric translation is acknowledged as a challenging field, requiring an interdisciplinary approach~\cite{susam2008translation}. As early as 1915, it was suggested that an ideal lyric translator should possess expertise in both linguistics and music, highlighting the need for a comprehensive understanding of the principles and techniques used in translation studies, coupled with a background in musicology~\cite{Spaeth1915}. Moreover, it is crucial to understand the cultural context of each language, such as different strategies employed for forming rhymes~\cite{low2008}. 
Because of these challenges, the systematic analysis and evaluation of lyric translation remain an under-researched topic of study. Thus far, only a few have proposed guidelines for scoring the quality of translated lyrics~\cite{low2005, low2008}.
While these principled approaches have proven successful, they lack automation. Consequently, despite the growing interest in the development of neural lyric translation, its evaluation has predominantly relied on human evaluation, making the evaluation process time-consuming, unreliable, and subjective~\cite{acl2022, acl2023, arxiv2023}.

Our study aims to computationally analyze and evaluate lyric translation based on a comprehensive understanding of lyrics that accounts for their musical, linguistic, and cultural elements. Unlike prior research that only proposed rhyme-scoring guidelines applicable to English~\cite{low2008}, our framework is extendable to Japanese and Korean. Though our framework may be limited in its application to specific languages, we strive to provide valuable insights into establishing distinct evaluation rules for phoneme repetition in diverse languages. Our comprehensive framework employs a multifaceted evaluation approach that examines lyric translation from four distinct perspectives: syllable counts, phoneme repetition, musical structure, and semantics. In the remainder of this paper, we explicate the rationale behind our selection of these perspectives by delving into the unique characteristics of lyric translation that differentiate it from general language translation tasks. In addition, we introduce the singable lyrics dataset we collected, which features line-by-line and section-by-section alignment of English, Japanese, and Korean lyrics. Moving forward, we propose robust evaluation metrics for lyric translation and analyze the results of our experiments based on the perspectives mentioned above. Finally, we conclude our paper by reflecting on the profound insights gleaned from our experiment and highlighting possible directions for future research. 

\section{Background}
Previous research indicates that linguistic analysis methods designed for standard text may not achieve desired outcomes when used to examine lyrics~\cite{lip}. Although automated evaluation metrics, such as $n$-gram-based~\cite{bleu, rouge, meteor} or neural approaches~\cite{bertscore}, have proven valuable and effective in assessing conventional machine translation tasks, they fall short in evaluating lyric translation. This is due to the unique characteristics of lyrics that render the translation process subject to many constraints and less direct~\cite{marc2015travelling}. 

One of the most significant constraints is the syllable count. This is because the original and translated lyrics must match the same melody lines, while it is a common practice to tweak the melody to accommodate minor changes in syllable count~\cite{low2008, hui2019translation}. In fact, conveying the same message in different languages requires vastly different syllable counts. For example, ``Happy New Year'' in English consists of 4 syllables, whereas 15 and 9 are required for Japanese (あけましておめでとうございます) and Korean (새해 복 많이 받으세요), respectively. For the numerical comparison, we examined PAWS-X, a dataset that contains 23,659 English sentences paired with human-translated sentences in various languages~\cite{paws2019}. The average number of syllables per sentence in the dataset is 50.89 for Japanese, whereas 36.18 per English and 40.40 per Korean. With these statistics, it can be deduced that Japanese necessitates approximately 41\% more syllables than English and 26\% more syllables than Korean to express an equivalent message. 
This limitation forces translators to often modify the meaning of original lyrics by adding, omitting, or even tweaking the message. However, translated lyrics still aim to capture the theme, mood, and spirit of the original lyrics~\cite{drinker1952}. Therefore, while original and translated lyrics need not be semantically identical, they still need to be semantically relevant~\cite{franzon2015three, low2008}.

It is also crucial to preserve the frequency of phoneme repetition (e.g., rhyme) in translated lyrics, particularly when the music demands it~\cite{drinker1952}. For instance, some sections, such as choruses, require a substantial degree of phoneme repetition, while others do not. Moreover, due to the inherent connection between lyrics and music, lyrics must be arranged in a way that complements the music~\cite{nlp_lyrics2005}. As a result, musically similar sections should maintain resembling linguistic features, including the choice of phonemes and the frequency of phoneme repetition~\cite{watanabe2020chorus}. 

\section{Dataset}
Although some websites provide user-translated multilingual lyrics, we found that most of them lack singability, as these translations were focused on delivering the meaning of the original lyrics rather than making them performable. While there are a few singable translations available, they are often not aligned on a line-by-line nor section-by-section basis due to the subjective nature of the lyric structure that there is no universal agreement on what to call a line and what to call a section. The absence of alignment makes it difficult to compare the original lyrics with their translated versions.
To address this issue, we collected a set of singable lyrics, sourced from either official lyrics of commercial songs or user-translated ones found on YouTube, meticulously aligned on a line-by-line basis in English, Japanese, and Korean. This approach ensures that lyrics on the same line share the same melodies. Moreover, the dataset divides the lyrics into sections, allowing for section-by-section analysis. Alongside the lyrics, it provides essential metadata such as genre, artist, original language, and the official status of lyrics. 
The dataset consists of 162 songs, each having lyrics in the three languages. It covers a diverse range of genres, including 109 K-pop, 23 animation music (e.g., Disney), 13 J-pop, 10 theatre music, and more. Table~\ref{tab:dataset_example} shows sample data.

\begin{table}[t]
\resizebox{\linewidth}{!}{%
\begin{tabular}{@{}lllll@{}}
\toprule
\textbf{\begin{tabular}[c]{@{}l@{}}Section\\ \#\end{tabular}} & \textbf{\begin{tabular}[c]{@{}l@{}}Line\\ \#\end{tabular}} & \textbf{English (EN)} & \textbf{Japanese (JP)} & \textbf{Korean (KR)} \\ \midrule
\multirow{6}{*}{1} & 1 & Twinkle, twinkle, little star & きらきらひかる & 반짝 반짝 작은별 \\
 & 2 & How I wonder what you are & おそらのほしよ & 아름답게 비치네 \\
 & 3 & Up above the world so high & まばたきしては & 서쪽 하늘에서도 \\
 & 4 & Like a diamond in the sky & みんなをみてる & 동쪽 하늘에서도 \\
 & 5 & Twinkle, twinkle, little star & きらきらひかる & 반짝 반짝 작은별 \\
 & 6 & How I wonder what you are & おそらのほしよ & 아름답게 비치네 \\ \midrule
\multirow{6}{*}{2} & 7 & Twinkle, twinkle, little star & きらきらひかる & 반짝 반짝 작은별 \\
 & 8 & How I wonder what you are & おそらのほしよ & 아름답게 비치네 \\
 & 9 & When the blazing sun is gone & みんなのうたが & 서쪽 하늘에서도 \\
 & 10 & When he nothing shines upon & とどくといいな & 동쪽 하늘에서도 \\
 & 11 & Then you show your little light & きらきらひかる & 반짝 반짝 작은별 \\
 & 12 & Twinkle, twinkle, all the night & おそらのほしよ & 아름답게 비치네 \\ \bottomrule
\end{tabular}}
\caption{Sample data illustrating the original English lyrics of ``Twinkle, Twinkle, Little Star'' and their corresponding singable translations in Japanese and Korean, aligned on a line-by-line and section-by-section basis.}
\label{tab:dataset_example}
\end{table}

\section{Evaluating Singability}
Our primary goal is to develop an evaluation framework that automatically assesses the quality of translated lyrics. One of the most important factors determining the quality is \textit{singability}, defined as not only the ability of being sung, but also the suitability (easiness) of being sung~\cite{franzon2015three}. To ensure such singability, we aim to provide metrics from three distinct perspectives by making sure that they i) maintain the song's melodic integrity, ii) preserve the degree of phoneme repetition, and iii) consider the underlying musical structure.

To substantiate the reliability of our evaluation metrics, we conducted a comparative analysis of singable lyrics versus non-singable lyrics based on each proposed evaluation metric. In all our comparative analyses, we utilized our dataset for singable lyrics, where official lyrics served as both source and target lyrics, and unofficial functioned as only target lyrics. For non-singable lyrics, we obtained pairs of original singable (source) and human-translated non-singable (target) lyrics, aligned line-wise and section-wise, for 3,642 songs from \url{https://lyricstranslate.com/}.

\begin{table}[b]
\centering
\resizebox{0.9\linewidth}{!}{%
\begin{tabular}{@{}ccclcl@{}}
\toprule
\multirow{1}{*}{\textbf{Source}} & \multirow{1}{*}{\textbf{Target}} & \multicolumn{2}{c}{\multirow{1}{*}{\textbf{Singable}}} & \multicolumn{2}{c}{\multirow{1}{*}{\textbf{Non-singable}}} \\ \midrule
\multirow{2}{*}{English} & Japanese & \multicolumn{2}{c}{0.17 (80 songs)} & \multicolumn{2}{c}{0.74 (1401 songs)} \\
 & Korean & \multicolumn{2}{c}{0.11 (80 songs)} & \multicolumn{2}{c}{0.48 (620 songs)} \\ \midrule
\multirow{2}{*}{Japanese} & English & \multicolumn{2}{c}{0.16 (162 songs)} & \multicolumn{2}{c}{0.39 (589 songs)} \\
 & Korean & \multicolumn{2}{c}{0.11 (162 songs)} & \multicolumn{2}{c}{0.31 (73 songs)} \\ \midrule
\multirow{2}{*}{Korean} & English & \multicolumn{2}{c}{0.09 (161 songs)} & \multicolumn{2}{c}{0.20 (702 songs)} \\
 & Japanese & \multicolumn{2}{c}{0.12 (161 songs)} & \multicolumn{2}{c}{0.52 (257 songs)} \\ \bottomrule
\end{tabular}
}
\caption{The average line syllable count distance ($Dis_{syl}$) between source and target languages for singable and non-singable lyrics.}
\label{tab:lscd}
\end{table}

\begin{table*}[t]
\centering
\resizebox{\linewidth}{!}{%
\begin{tabular}{@{}lllll@{}}
\toprule
\textbf{Section} & \textbf{English} & \textbf{Japanese (English translation)} & \textbf{Korean (English translation)} & \textbf{$pho$} \\ \midrule
\begin{tabular}[c]{@{}l@{}}$E(A_1)$,\\ $J(A_1)$,\\ $K(A_1)$\end{tabular} & \begin{tabular}[c]{@{}l@{}}Do you wanna build a snowman?\\ Come on, let's go and play!\\ I never see you anymore\\ Come out the door\\ It's like you've gone away\end{tabular} & \begin{tabular}[c]{@{}l@{}}\begin{CJK}{UTF8}{min}雪だるま作ろう\end{CJK} (Let’s build a snowman)\\ \begin{CJK}{UTF8}{min}ドアを開けて\end{CJK} (Please open the door)\\ \begin{CJK}{UTF8}{min}一緒に遊ぼう\end{CJK} (Let’s play together)\\ \begin{CJK}{UTF8}{min}どうして\end{CJK} (Why)\\ \begin{CJK}{UTF8}{min}出てこないの?\end{CJK} (don’t you come out?)\end{tabular} & \begin{tabular}[c]{@{}l@{}}같이 눈사람 만들래? (Do you wanna build a snowman?)\\ 제발 좀 나와봐 (Please come out)\\ 언니를 만날 수 없어 (I can’t meet you)\\ 같이 놀자 (Let’s play together)\\ 나 혼자 심심해 (I’m lonely alone)\end{tabular} & \begin{tabular}[c]{@{}l@{}}0.85,\\ 0.73, \\ 0.77\end{tabular} \\ \midrule
\begin{tabular}[c]{@{}l@{}}$E(B_1)$,\\ $J(B_1)$,\\ $K(B_1)$\end{tabular} & \begin{tabular}[c]{@{}l@{}}We used to be best buddies\\ And now we're not\\ I wish you would tell me why!\end{tabular} & \begin{tabular}[c]{@{}l@{}}\begin{CJK}{UTF8}{min}前は仲良く\end{CJK}(We were close before)\\ \begin{CJK}{UTF8}{min}してたのに\end{CJK} (We used to be)\\ \begin{CJK}{UTF8}{min}なぜ会えないの\end{CJK} (Why can’t we meet each other?)\end{tabular} & \begin{tabular}[c]{@{}l@{}}그렇게 친했는데 (We were close before)\\ 이젠 아냐 (and we’re not)\\ 그 이유를 알고파 (I want to know the reason why)\end{tabular} & \begin{tabular}[c]{@{}l@{}}0.92, \\ 0.80, \\ 0.91\end{tabular} \\ \midrule
\begin{tabular}[c]{@{}l@{}}$E(A_2)$,\\ $J(A_2)$,\\ $K(A_2)$\end{tabular} & \begin{tabular}[c]{@{}l@{}}Do you wanna build a snowman?\\ Or ride our bike around the halls?\\ I think some company is overdue\\ I've started talking to the pictures on the walls!\end{tabular} & \begin{tabular}[c]{@{}l@{}}\begin{CJK}{UTF8}{min}雪だるま作ろう\end{CJK} (Let’s build a snowman)\\ \begin{CJK}{UTF8}{min}自転車に乗ろう\end{CJK}(Let’s ride a bike)\\ \begin{CJK}{UTF8}{min}ずっと一人でいると\end{CJK} (When I'm alone all the time)\\ \begin{CJK}{UTF8}{min}壁の絵とおしゃべりしちゃう\end{CJK}\\ (I’m almost talking to the pictures on the walls)\end{tabular} & \begin{tabular}[c]{@{}l@{}}같이 눈사람 만들래? (Do you wanna build a snowman?)\\ 아니면 자전거 탈래? (or do you wanna ride a bike?)\\ 이제는 나도 지쳐 가나봐 (Seems I’m getting tired)\\ 벽에다 말을 하며 놀고 있잖아\\ (because I’ve started talking to the walls)\end{tabular} & \begin{tabular}[c]{@{}l@{}}0.79, \\ 0.73, \\ 0.82\end{tabular} \\ \midrule
\begin{tabular}[c]{@{}l@{}}$E(B_2)$,\\ $J(B_2)$,\\ $K(B_2)$\end{tabular} & \begin{tabular}[c]{@{}l@{}}It gets a little lonely\\ All these empty rooms\\ Just watching the hours tick by\end{tabular} & \begin{tabular}[c]{@{}l@{}}\begin{CJK}{UTF8}{min}さびしい部屋で\end{CJK} (In a lonely room)\\ \begin{CJK}{UTF8}{min}柱時計\end{CJK} (the wall clock)\\ \begin{CJK}{UTF8}{min}見てたりするの\end{CJK} (I look at or something)\end{tabular} & \begin{tabular}[c]{@{}l@{}}사실은 조금 외로워 (In fact, I’m a little lonely)\\ 텅빈 방에선 (In empty rooms)\\ 시계소리만 들려 (All I can hear is the clock's ticking)\end{tabular} & \begin{tabular}[c]{@{}l@{}}0.90, \\ 0.88, \\ 0.96\end{tabular} \\ \bottomrule
\end{tabular}
}
\caption{Lyric excerpt from ``Do You Want to Build a Snowman'' from the animation ``Frozen,'' singable in all languages. Sections $A_1$ and $A_2$ form a musically similar pair, while $B_1$ and $B_2$ are also musically similar to each other. Each section is denoted as $E(A_1), \dots ,E(B_2)$ in English, $J(A_1), \dots ,J(B_2)$ in Japanese, and $K(A_1), \dots ,K(B_2)$ in Korean.}
\label{tab:snowman}
\vspace*{-.9mm}
\end{table*}

\subsection{Line Syllable Count Distance}
It is crucial to preserve the syllable counts between the original and translated lyrics for each line as similar as possible in order to maintain the integrity of a song's melody~\cite{apter2012translating}. Therefore, it is unsurprising that our evaluation framework incorporates a metric to assess the differences in syllable counts. Let the line syllable counts for a pair of lyrics that consist of $n$ lines, $\mathbf{X}=\{{x_1}, ..., {x_n}\}$ and $\mathbf{\Tilde{X}}=\{\Tilde{x_1}, ..., \Tilde{x_n}\}$ be denoted as $\{syl(x_1), ..., syl(x_n)\}$ and $\{syl(\Tilde{x_1}), ..., syl(\Tilde{x_n})\}$ where each element refers to the syllable count of each line. For instance, if the first line of the English lyrics $\mathbf{X}$ is ``Silent night holy night'' and the corresponding line in the Korean lyrics $\mathbf{\Tilde{X}}$ is ``Goyohanbam-georukhanbam (고요한밤 거룩한밤)'', the value of $syl(x_1)$ is 6 and $syl(\Tilde{x_1})$ is 8. We define the \textbf{line syllable count distance} between a pair of lyrics $\mathbf{X}$ and $\mathbf{\Tilde{X}}$ ($Dis_{syl}{(\mathbf{X},\mathbf{\Tilde{X}})}$) in order to evaluate the disparities in syllable counts, as follows.
\vspace*{-2mm}
\begin{equation}
\begin{aligned}
\scalebox{0.93}{$
{Dis_{syl}{(\mathbf{X},\mathbf{\Tilde{X}})}=\frac{1}{2n}\sum_{i=1}^{n} (\frac{|{syl(x_i)} - {syl(\Tilde{x_i})}|}{syl(x_i)}+} {\frac{|{syl(x_i)} - {syl(\Tilde{x_i})}|}{syl(\Tilde{x_i})})}
$}
\end{aligned}
\end{equation}
We compare the line syllable count distance of singable and non-singable lyrics. As shown in Table \ref{tab:lscd}, non-singable lyrics display a considerably greater $Dis_{syl}{(\mathbf{X}, \mathbf{\Tilde{X}})}$ compared to singable lyrics due to the varying syllable count requirements across languages.

\subsection{Phoneme Repetition Similarity}
Rhyme, defined as the repetition of a vowel sound and any subsequent sounds~\cite{bain1867english}, has historically been a fundamental element in the realm of poetry, including in Western languages like English. However, the concept of rhyme has not been as prevalent in Japanese or Korean poetry~\cite{japan_hiphop}. In fact, traditional Korean poetry did not incorporate this concept~\cite{yoon2001language}. Despite the increasing tendency to adopt the concept of rhyme in Japanese and Korean lyrics due to intercultural exchanges, we observed that lyrics in these languages often rely more on repeating grammatical elements. 
For example, in section $A_1$ of Table~\ref{tab:snowman}, the Japanese pair ``tsukurou (作ろう, Let's build)'' and ``asobou (遊ぼう, Let's play)'' generates a sense of rhyme because both end with the same conjugation ``ou'' meaning ``let's''. Similarly, in Section $A_2$, the Korean pair ``mandeullae (만들래, Do you wanna build)'' and ``tallae (탈래, Do you wanna ride)'' creates a sense of repetition because both end with ``llae'' meaning ``Do you wanna''. Another example is the repetition of particles at the end of sentences, such as ``yo (よ)'' and ``no (の)'' in Japanese and ``yo (요)'' and ``da (다)'' in Korean, which convey cultural nuances related to formality. We therefore propose that English, Japanese, and Korean share common ground in adopting phoneme repetition for poetic expression. However, as such repetition is not necessarily called rhyme in Japanese and Korean, we will refrain from using the term ``rhyme'' and instead employ the term ``phoneme repetition.''

We noticed that each section's degree of phoneme repetition remains consistent across different languages when the lyrics are singable. For example, in Table~\ref{tab:snowman}, the first section of the original English lyrics ($E(A_1)$) displays a strong degree of phoneme repetition, with three rhyming pairs: ``come-come'', ``play-away'', and ``anymore-door'' (In this paper, we denote a section as an uppercase with a number and a line as a lower case with a number). Similarly, both the Japanese and Korean translations ($J(A_1)$, $K(A_1)$) also exhibit a substantial degree of phoneme repetition, featuring three pairs of repeated phonemes in each: ``doa (ドア)''-``dou (どう)'', ``tsukurou (作ろう)''-``asobou (遊ぼう)'', ``akete (開けて)''-``shite (して)'' in Japanese, and ``gachi (같이)''-``gachi (같이)'', ``mandeul (만들)''-``eonnireul (언니를)'', ``mandeullae (만들래)''-``simsimhae (심심해)'' in Korean. However, we realized that it is not fair to directly compare the number of phoneme repetitions when attempting to quantify the degree of phoneme repetition as each language has a different number of vowels and consonants: English has 15 vowels and 24 consonants, whereas Japanese has 5 and 15 and Korean has 21 and 19. Hence, in an attempt to minimize the differences in the number of phonemes, we treated acoustically similar vowels as the same vowel in English, such as \lq IH\rq-\lq IY\rq, \lq UH\rq-\lq UW\rq, or \lq EH\rq-\lq AE\rq  (e.g., 'mass' and 'mess')~\cite{syrdal1986perceptual} because they can still form slant rhymes~\cite{hanson2003formal}. Conversely, we considered \lq A\rq-\lq YA\rq, \lq O\rq-\lq YO\rq, and \lq U\rq-\lq YU' as separate vowels in Japanese, as they are unlikely to function as the same grammatical components. In Korean, we regarded the perceptually similar vowels (e.g., \lq AE\rq-\lq E\rq or \lq OE\rq-\lq OI\rq-\lq OAE\rq) as the same vowels~\cite{ito2006adaptation, chang2017}.

To quantitatively represent the degree of phoneme repetition, we utilized the concept of \emph{distinct-2}, the ratio of the number of distinct bi-grams to the total number of bi-grams~\cite{distinct_n}. While the original concept formed bi-grams using two consecutive words, we used two consecutive phonemes to assess the degree of repetition because lower \emph{distinct-2} values indicate higher repetition and vice versa. The \emph{phoneme distinct-2} ($pho$) of a section $X_i$, is defined as follows{:} 
\vspace*{-2mm}
\begin{equation}
\begin{aligned}
\scalebox{0.90}{$
pho(X_i) = \frac{\mbox{\it unique\;bi-gram\;\#\;in}\;X_i}{\mbox{\it total\;bi-gram\;\#\;in}\;X_i}$} .
\end{aligned}
\end{equation}
For example, consider a section with a single line ``twinkle twinkle little star'', denoted as $X_1$. First, we decomposed the section into phonemes and added the \lq \texttt{\textless{}eos\textgreater}\rq to each line: \lq T\rq, \lq W\rq, \lq IH\rq, \lq NG\rq, \lq K\rq, \lq AH\rq, \dots, \lq S\rq, \lq T\rq, \lq AA\rq, \lq R\rq, and \lq\texttt{\textless{}eos\textgreater}\rq. Next, we grouped each component into bi-grams: \lq TW\rq, \lq WIH\rq, \lq IHNG\rq, \lq NGK\rq, \lq KAH\rq, \lq AHL\rq, \dots, \lq ST\rq, \lq TAA\rq, \lq AAR\rq, \lq R\texttt{\textless{}eos\textgreater}\rq. Finally, we calculated the \emph{phoneme distinct-2} of the section ($pho(X_1)$) by dividing the number of unique bi-grams by the total number of bi-grams (17/23 = 0.74).
To measure the similarity between two sections in terms of the degree of phoneme repetition, we introduce the \textbf{phoneme repetition similarity} ($Sim_{pho}$). Given two sets of lyrics with $m$ sections, $\mathbf{X}=\{X_1,...,X_m\}$ and $\mathbf{\Tilde{X}}=\{\Tilde{X_1},...,\Tilde{X_m}\}$, the phoneme repetition similarity between $\mathbf{X}$ and $\mathbf{\Tilde{X}}$ is defined as the Spearman correlation between $\{pho(X_1),...,pho(X_m)\}$ and $\{pho(\Tilde{X_1}),...,pho(\Tilde{X_m)\}}$, as shown below. 
\vspace{-1mm}
\begin{equation}
\begin{aligned}
\scalebox{0.75}{$
Sim_{pho}(\mathbf{X},\mathbf{\Tilde{X}}) = corr(\{pho(X_1),...,pho(X_m)\},
\{pho(\Tilde{X_1}),...,pho(\Tilde{X_m})\})
$}
\end{aligned}
\end{equation}
We present the statistical results for the average phoneme repetition similarity of singable and non-singable lyrics in Table~\ref{tab:phoneme}. The table clearly exhibits a higher correlation between the original lyrics and singable translated lyrics in terms of the \emph{phoneme distinct-2} than non-singable lyrics. This result suggests that singable lyric translation takes into account the degree of phoneme repetition within each section to convey a sense of repetition for that section. 

\begin{table}[t]
\centering
\resizebox{0.72\linewidth}{!}{%
\begin{tabular}{@{}ccclcl@{}}
\toprule
\multirow{1}{*}{\textbf{Source}} & \multirow{1}{*}{\textbf{Target}} & \multicolumn{2}{c}{\multirow{1}{*}{\textbf{Singable}}} & \multicolumn{2}{c}{\multirow{1}{*}{\textbf{Non-singable}}} \\ \midrule
\multirow{2}{*}{English} & Japanese & \multicolumn{2}{c}{0.69} & \multicolumn{2}{c}{0.56} \\
 & Korean & \multicolumn{2}{c}{0.97} & \multicolumn{2}{c}{0.72} \\ \midrule
\multirow{2}{*}{Japanese} & English & \multicolumn{2}{c}{0.79} & \multicolumn{2}{c}{0.61} \\
 & Korean & \multicolumn{2}{c}{0.80} & \multicolumn{2}{c}{0.48} \\ \midrule
\multirow{2}{*}{Korean} & English & \multicolumn{2}{c}{0.97} & \multicolumn{2}{c}{0.78} \\
 & Japanese & \multicolumn{2}{c}{0.80} & \multicolumn{2}{c}{0.50} \\ \bottomrule
\end{tabular}
}
\caption{The average phoneme repetition similarity ($Sim_{pho}$) of singable lyrics and non-singable lyrics.}
\label{tab:phoneme}
\end{table}

\begin{figure}[t]
 \centerline{
 \includegraphics[width=0.9\linewidth]{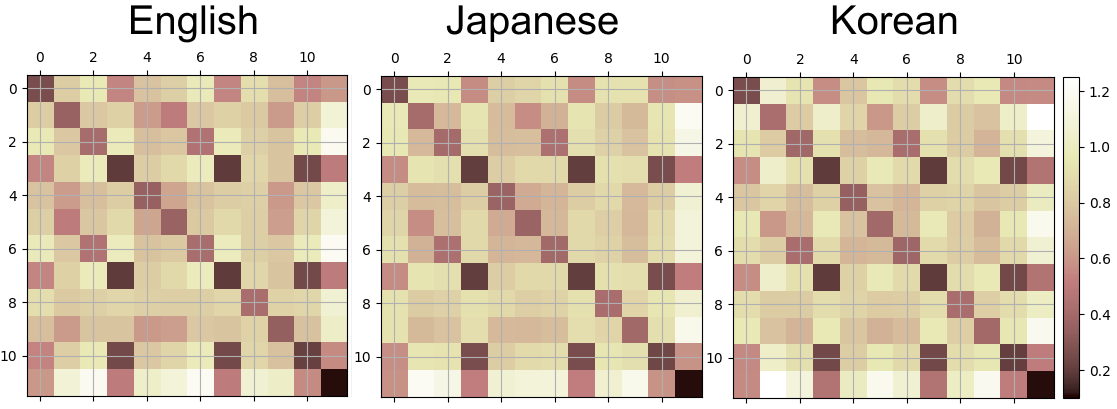}}
\vspace*{-2mm}
\caption{Musical self-dissimilarity matrices for English, Japanese, and Korean versions of the K-pop song ``Icy'' by ITZY. Dissimilarity between the $i$-th and the $j$-th section was computed using $diss(X_i, X_j)$.}
 \label{fig:self_matrix}
\end{figure}

\subsection{Musical Structure Distance}
Upon examining our section-divided singable lyrics data, we identified two tendencies in lyrics when musical sections are repeated (e.g., the repetition of the chorus). First, we observed that musically similar sections tend to share the same phonemes and, as expected, the same phrases. For instance, in Table~\ref{tab:snowman}, musically similar sections share the same vowels (e.g., ``why'' in $E(B_1)$ and ``by'' in $E(B_2)$) or identical phrases (e.g., ``Do you wanna build a snowman'' in $E(A_1)$ and $E(A_2)$) in order to create a sense of consistency. As a result, when calculating the \emph{phoneme distinct-2} ($pho$) for two concatenated sections, musically similar sections are likely to have smaller values than musically different sections. For example, $pho(E({A_1}\concat{A_2}))$ is 0.70 (`$\concat$' denotes the concatenation of text), whereas $pho(E({A_1}\concat{B_1}))$ is 0.82, where $A_1$ is musically similar to $A_2$ but not to $B_1$. However, a low value of $pho$ does not always imply musical similarity, as a meager $pho$ value in one section could result in a low $pho$ of two concatenated sections despite the musical dissimilarity (e.g., ``lalalalalalalalalalalalalala'' $\concat$ ``do you wanna build a snowman?''). From this case, we derived our second observation that a significant difference in $pho$ for each section could imply musical differences. Accordingly, we also realized that musically similar sections tend to have a similar degree of $pho$. For instance, in Table~\ref{tab:snowman}, both $A_1$ and $A_2$, a set of musically similar sections, exhibit relatively low $pho$, indicating a strong degree of phoneme repetition (rhyme), with similar values to each other across all languages. Likewise, both $B_1$ and $B_2$, another pair of musically similar sections, demonstrate a higher $pho$, with similar values to each, in all languages.

Therefore, to quantify the musical similarity between sections, we examined whether they have 1) a tendency to share the same phoneme by obtaining $pho(X_i{\concat}X_j)$, and 2) similar $pho$ values by calculating $|pho(X_i)-pho(X_j)|$. Given that higher values represent dissimilarity in both cases, we define the musical dissimilarity between two sections, $diss(X_i,X_j)$, as the sum of these two values, as follows.
\vspace{-1mm}
\begin{equation}
\begin{aligned}
\scalebox{0.9}{$diss(X_i,X_j)=pho(X_i{\concat}X_j)+|pho(X_i) - pho(X_j)|$}
\end{aligned}
\end{equation}
As shown in Figure~\ref{fig:self_matrix}, self-dissimilarity matrices employing our definition of musical dissimilarity look highly similar across English, Japanese, and Korean, where all are singable, visually representing musical structure.

Finally, we quantitatively evaluated the distance between matrices. We refer to this distance between matrices as the musical structure distance, as it represents the structural element of lyrics. In summary, the \textbf{musical structure distance} between lyrics in different languages $\mathbf{X}$ and $\mathbf{\Tilde{X}}$, each consisting of $m$ sections, $Dis_{mus}(\mathbf{X}, \mathbf{\Tilde{X}})$, is defined as follows:
\vspace{-2mm}
\begin{equation}
\begin{aligned}
\scalebox{0.85}{$Dis_{mus}(\mathbf{X},\mathbf{\Tilde{X}})= \frac{1}{m^2}  
\sqrt{\sum_{i, j=1}^{m}{(diss(X_i,X_j)-diss(\Tilde{X_i},\Tilde{X_j})})^2}$}
\end{aligned}
\end{equation}
In Table \ref{tab:structure}, we provide a summary of the average musical structure distance for singable lyrics, human-translated non-singable lyrics, and machine-translated non-singable lyrics generated by automatically translating official singable lyrics from 80 English, 162 Japanese, and 161 Korean songs using Google Translator. Our findings show that singable lyrics exhibit the lowest $Dis_{mus}$ values, while machine-translated non-singable lyrics display the highest, suggesting that machine-translated ones lack structural coherence the most. As human-translated non-singable lyrics maintain structural coherence in aspects such as word choice and nuances, they demonstrate lower distances than machine-translated counterparts.

\begin{table}[t]
\centering
\resizebox{0.96\linewidth}{!}{%
\begin{tabular}{@{}ccccc@{}}
\toprule
\textbf{Source} & \textbf{Target} & \multicolumn{1}{c}{\textbf{Singable}} & \multicolumn{1}{c}{\textbf{\begin{tabular}[c]{@{}c@{}}Non-singable\\ (Human)\end{tabular}}} & \multicolumn{1}{c}{\textbf{\begin{tabular}[c]{@{}c@{}}Non-singable\\ (Machine)\end{tabular}}} \\ \midrule
\multirow{2}{*}{English} & Japanese & 0.14 & 0.26 & 0.30 \\
 & Korean & 0.10 & 0.15 & 0.15 \\ \midrule
\multirow{2}{*}{Japanese} & English & 0.13 & 0.20 & 0.18 \\
 & Korean & 0.11 & 0.13 & 0.14 \\ \midrule
\multirow{2}{*}{Korean} & English & 0.10 & 0.10 & 0.10 \\
 & Japanese & 0.11 & 0.10 & 0.17 \\ \bottomrule
\end{tabular}
}
\caption{The average musical structure distance ($Dis_{mus}$) of singable lyrics and non-singable lyrics.}
\label{tab:structure}
\vspace{-1.6mm}
\end{table}

\section{Evaluating Semantics}
Semantic relatedness to the original lyrics is by no means less fundamental than syllable counts, phoneme repetition, and structural factors~\cite{franzon2015three}. We therefore introduce a fourth metric, semantic similarity, to ensure the semantic relevance of translated lyrics to the original.

\subsection{Semantic Similarity}
To numerically assess the semantic textual similarity ($sts$) between a pair of lyrics, we first obtained the contextual embeddings of each text from lyrics using a pre-trained Sentence BERT model \footnote{We used \texttt{all-MiniLM-L6-v2} \cite{minilm2020}.} \cite{reimer_2019} and then calculated the cosine similarity between the embeddings. 
As this model was trained for English, the Japanese and Korean lyrics were automatically translated using Google Translator before obtaining the embeddings. 

\begin{figure}[t]
\centerline{
\includegraphics[width=0.89\linewidth]{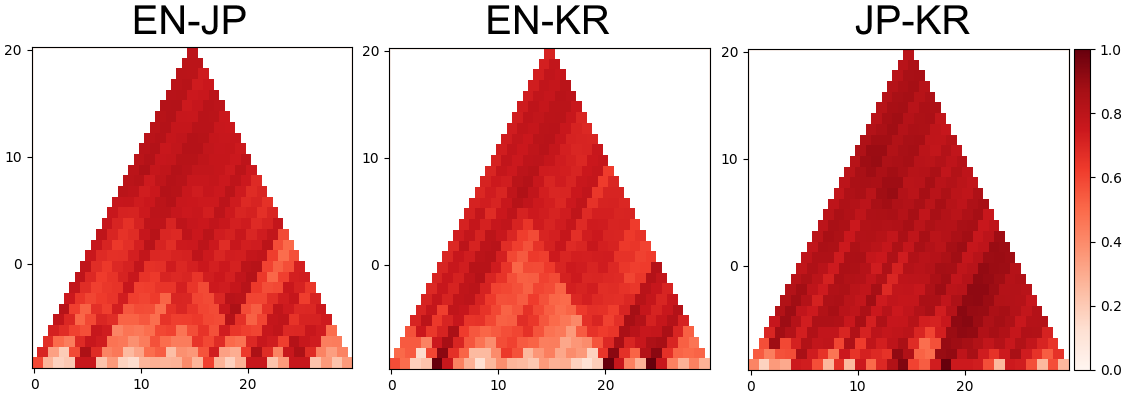}}
\vspace*{-3mm}
\caption{Semantic similarity cross-scape plot for the J-pop song ``A Thousand Winds'' between English and Japanese (\textbf{Left}), English and Korean (\textbf{Middle}), and Japanese and Korean (\textbf{Right}). Any value less than 0 was regarded as 0.}
\label{fig:crossplot}
\end{figure}

We started by examining hierarchical semantic similarity using cross-scape plots~\cite{park2019}, as shown in Figure~\ref{fig:crossplot}. Given a pair of lyrics $\mathbf{X} =\{x_1, \dots, x_n\}$  and $\mathbf{\Tilde{X}}= \{\Tilde{x_1}, \dots, \Tilde{x_n}\}$ with $n$ lines each, the first (leftmost) block of the lowest line represents the semantic textual similarity between $x_1$ and $\Tilde{x_1}$ (denoted as $sts(x_1, \Tilde{x_1})$) while the last (rightmost) block signifies $sts(x_n, \Tilde{x_n})$. The first (leftmost) block of the second-lowest line denotes $sts({x_1{\concat}x_2}, {\Tilde{x_1}{\concat}\Tilde{x_2}})$, and the second block corresponds to $sts({x_2{\concat}x_3}, {\Tilde{x_2}{\concat}\Tilde{x_3}})$. Lastly, the highest block (line) represents the similarity between the entire lyrics, $sts(x_1\concat \dots \concat x_n, \Tilde{x_1} \concat \dots \concat \Tilde{x_n})$.

\begin{table}[t]
\centering
\resizebox{0.90\linewidth}{!}{%
\begin{tabular}{@{}lllc@{}}
\toprule
{\color[HTML]{453123} \textbf{\begin{tabular}[c]{@{}l@{}}Line\\ \#\end{tabular}}} & {\color[HTML]{453123} \textbf{English}} & \textbf{\begin{tabular}[c]{@{}l@{}}Japanese\\ (English translation)\end{tabular}} & {\color[HTML]{453123} \textbf{\begin{tabular}[c]{@{}l@{}}$sts$\end{tabular}}} \\ \midrule
{\color[HTML]{453123} 1} & \cellcolor[HTML]{00FFFF}\begin{tabular}[c]{@{}l@{}}please do not stand at \\ my grave and weep.\end{tabular} & \cellcolor[HTML]{00FFFF}\begin{tabular}[c]{@{}l@{}}私のお墓の前で\\ (in front of my grave)\end{tabular} & {\color[HTML]{453123} 0.56} \\ \midrule
{\color[HTML]{453123} 2} & \cellcolor[HTML]{FF00FF}\begin{tabular}[c]{@{}l@{}}I am not there, I do not sleep\end{tabular} & \cellcolor[HTML]{00FFFF}\begin{tabular}[c]{@{}l@{}}泣かないでください\\ (please stop crying)\end{tabular} & {\color[HTML]{453123} 0.27} \\ \midrule
{\color[HTML]{453123} \begin{tabular}[c]{@{}l@{}}1,\\ 2\end{tabular}} & \begin{tabular}[c]{@{}l@{}}please do not stand at\\ my grave and weep. I am\\ not there, I do not sleep\end{tabular} & \begin{tabular}[c]{@{}l@{}}私のお墓の前で\\ 泣かないでください\\ (in front of my grave.\\ please stop crying.)\end{tabular} & {\color[HTML]{453123} {\ul \textbf{0.76}}} \\ \bottomrule
\end{tabular}
}
\caption{Semantic textual similarity ($sts$) between English and Japanese versions of ``A Thousand Winds''.}
\label{tab:thousand_winds}
\vspace*{-2mm}
\end{table}

In each plot of Figure~\ref{fig:crossplot}, there are semantic disparities at lower levels, but similarities increase at higher (broader) levels. We have two explanations for this. First, the number of musical notes within a single lyric line may be adequate to deliver a specific message in one language but insufficient in another language. Therefore, it is common for a message conveyed in one line in one language to span two lines in another language. As an example, we provide Table~\ref{tab:thousand_winds}, which presents the semantic textual similarity ($sts$) between Japanese and English lyrics of the J-pop song ``A Thousand Winds (千の風になって)''. As demonstrated in the table, the similarity between Japanese and English at a broader level ($sts({x_1{\concat}x_2}, {\Tilde{x_1}{\concat}\Tilde{x_2}})$) can be higher than at the line level ($sts(x_1,\Tilde{x_1})$, $sts(x_2,\Tilde{x_2})$) because Japanese generally requires more syllables than English and it often takes two lines in Japanese to express a single-line message in English. 
Second, the semantic similarities at broader levels can be higher because of grammatical/linguistic differences. Each language has its own natural word order patterns. For example, in the phrase ``I'm going to travel to find the gold,'' it is natural in English to mention ``I'm going to travel'' before ``to find the gold.'' However, in Japanese and Korean, expressing ``to find the gold (金を探しに, 금을 찾으러)'' before ``I'm going to travel (旅に出る, 떠난다)'' is a more typical and natural construction. Table~\ref{tab:gold} shows that these differences between languages make line-level semantic assessments insufficient. Since lines 1, 2, and 3 in the English version correspond to lines 3, 1, and 2 respectively in the Japanese version, these pairs exhibit low semantic similarities at the line level ($sts(x_1,\Tilde{x_1})$, $sts(x_2,\Tilde{x_2})$, $sts(x_3,\Tilde{x_3})$), while demonstrating higher similarity when considered as a whole ($sts(x_1{\concat}x_2{\concat}x_3,\Tilde{x_1}{\concat}\Tilde{x_2}{\concat}{x_3})$). 

\begin{table}[t]
\resizebox{\linewidth}{!}{%
\begin{tabular}{@{}lllc@{}}
\toprule
\textbf{\begin{tabular}[c]{@{}l@{}}Line\\ \#\end{tabular}} & \textbf{English} & \textbf{\begin{tabular}[c]{@{}l@{}}Japanese\\ (English translation)\end{tabular}} & \textbf{\begin{tabular}[c]{@{}l@{}}$sts$\end{tabular}} \\ \midrule
1 & \cellcolor[HTML]{CFE2F3}{\color[HTML]{150E0C} \begin{tabular}[c]{@{}l@{}}Dare to try and reach out \\ for hеaven\end{tabular}} & \cellcolor[HTML]{FFF2CC}{\color[HTML]{150E0C} \begin{tabular}[c]{@{}l@{}}望むように生きるなら (If you want \\to become what you’re meant to be)\end{tabular}} & 0.22 \\ \midrule
2 & \cellcolor[HTML]{FFF2CC}{\color[HTML]{150E0C} \begin{tabular}[c]{@{}l@{}}You must become \\what you'rе meant to be\end{tabular}} & \cellcolor[HTML]{00FF00}{\color[HTML]{150E0C} \begin{tabular}[c]{@{}l@{}}星からの金を求め\\(to find the gold from stars)\end{tabular}} & 0.27 \\ \midrule
3 & \cellcolor[HTML]{00FF00}{\color[HTML]{150E0C} \begin{tabular}[c]{@{}l@{}}And bring the gold of \\heaven to the world\end{tabular}} & \cellcolor[HTML]{C9DAF8}{\color[HTML]{150E0C} \begin{tabular}[c]{@{}l@{}}一人旅に出るのよ\\(Dare to embark on a solo journey)\end{tabular}} & 0.13 \\ \midrule
1-3& {\color[HTML]{150E0C} \begin{tabular}[c]{@{}l@{}}Dare to try and reach out \\for hеaven You must become \\what you'rе meant to be\\And bring the gold of \\heaven to the world\end{tabular}} & {\color[HTML]{150E0C} \begin{tabular}[c]{@{}l@{}}望むように生きるなら星からの\\金を求め一人旅に出るのよ\\(Dare to embark on a solo journey\\if you want to become what you’re\\meant to be to find the gold from stars)\end{tabular}\hspace*{-3mm}} & {\ul \textbf{0.53}} \\ \bottomrule
\end{tabular}
}
\caption{Semantic textual similarity ($sts$) between English and Japanese versions of ``Gold von den Sternen''.}
\label{tab:gold}
\end{table}

\begin{figure}[t]
 \centerline{
 \includegraphics[width=0.76\linewidth]{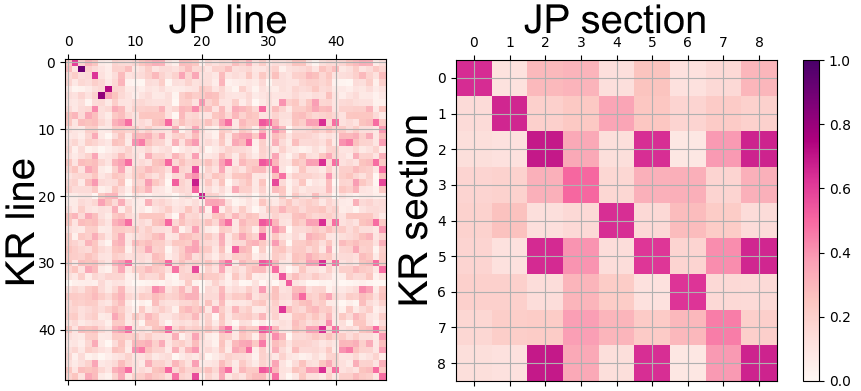}}
\vspace*{-2mm}
\caption{The line-wise (\textbf{Left}) and section-wise (\textbf{Right}) semantic similarity matrices between Japanese and Korean versions of ``Wie wird man seinen Schatten los?''}
 \label{fig:cross_semantics}
\end{figure}

Considering these factors, it becomes evident that singable lyric translations do not prioritize line-wise semantic similarity. Rather, we observed that singable translations aim to preserve semantic connections at the section level since the organization of the lyric storyline follows a section-wise approach. To illustrate this, we present Figure~\ref{fig:cross_semantics}, which displays both line-wise and section-wise semantic similarity matrices for the Japanese and Korean versions of ``How do you get rid of your shadow? (Wie wird man seinen Schatten los?)'' from the German musical ``Mozart!''. As shown in the Figure, the section-wise matrix represents the semantic relatedness more clearly than the line-wise matrix. 

Therefore, we propose assessing section-wise semantic relatedness for evaluating singable lyric translation. To achieve this, we define the \textbf{semantic similarity} between a pair of lyrics $\mathbf{X}= \{X_1,...,X_m\}$ and $\mathbf{\Tilde{X}}=\{\Tilde{X_1},...,\Tilde{X_m}\}$, consisting of $m$ sections and $n = n(X_1) + \dots + n(X_m)$ lines, where $n(X_i)$ denotes the number of lines in the $i$-th section, as follows:
\vspace*{-2mm}
\begin{equation}
\begin{aligned}
\scalebox{0.90}{$Sim_{sem}(\mathbf{X},\mathbf{\Tilde{X}})=\sum_{i=1}^{m} (\frac{n(X_i)}{n}sts(X_i, \Tilde{X_i}))$} .
\end{aligned}
\end{equation}
Table~\ref{tab:semantic} compares singable and non-singable lyrics in terms of line-wise semantic similarities ($\frac{1}{n}\sum_{i=1}^{n} sts(x_i, \Tilde{x_i})$) and section-wise similarities, using our proposed metric ($Sim_{sem}$). The table reveals that non-singable translations exhibit high semantic similarity for both line-wise and section-wise measures, with similar values for each. In contrast, singable translations show low line-wise similarity, as expected, since they do not prioritize line-wise semantic similarity. However, when evaluated using section-wise similarity, they display a level of similarity comparable to that between ``Machine learning is so easy'' and ``Deep learning is so straightforward'', which is 0.623 when measured with the same pre-trained model~\cite{minilm2020}.

\begin{table}[t]
\centering
\resizebox{0.79\linewidth}{!}{%
\begin{tabular}{@{}cccccc@{}}
\toprule
\multirow{2}{*}{\textbf{Source}} & \multirow{2}{*}{\textbf{Target}} & \multicolumn{2}{c}{\textbf{Singable}} & \multicolumn{2}{c}{\textbf{Non-singable}} \\ \cmidrule(l){3-6} 
 &  & \textbf{line} & \textbf{section} & \textbf{line} & \textbf{section} \\ \midrule
\multirow{2}{*}{English} & Japanese & 0.40 & 0.54 & 0.64 & 0.74 \\
 & Korean & 0.42 & 0.55 & 0.70 & 0.76 \\ \midrule
\multirow{2}{*}{Japanese} & English & 0.47 & 0.59 & 0.66 & 0.72 \\
 & Korean & 0.52 & 0.61 & 0.77 & 0.79 \\ \midrule
\multirow{2}{*}{Korean} & English & 0.53 & 0.63 & 0.78 & 0.81 \\
 & Japanese & 0.52 & 0.61 & 0.73 & 0.75 \\ \bottomrule
\end{tabular}
}
\caption{The average line-wise semantic similarity and section-wise semantic similarity ($Sim_{sem}$) of singable and non-singable lyrics.}
\label{tab:semantic}
\end{table}

\section{Discussions and Conclusions}
In this paper, we introduced a computational evaluation framework for singable lyric translation, grounded in the musical, linguistic, and cultural understanding of lyrics, comprised of four evaluation metrics, line syllable count distance ($Dis_{syl}$), phoneme repetition similarity ($Sim_{pho}$), musical structure distance ($Dis_{mus}$), and semantic similarity ($Sim_{sem}$). These metrics are designed to ensure that the translated lyrics maintain the integrity of melodies, degree of phoneme repetition, structural coherence, and semantics of the original lyrics. Our framework is automated, guaranteeing objectivity and efficiency in terms of time and cost. We showed the efficacy of our evaluation metrics by offering comparative statistics between singable and non-singable lyrics. In addition, our analysis revealed that the degree of phoneme repetition in the original lyrics is frequently mirrored in the translated lyrics, musically similar sections tend to share the same phonemes and display comparable degrees of phoneme repetition, and section-wise analysis is better suited for evaluating semantic similarity for lyric translation than line-wise analysis.

Nonetheless, there remains room for improvement. 
Although we have assembled a singable lyrics dataset, aligned across English, Japanese, and Korean, our dataset has some limitations; it lacks musical information and its volume is limited. As a result, we have not been able to incorporate musical notes into our experiment or conduct comparative studies across various genres. We recognize that an ideal lyric translation evaluation system should take into account the relationship between musical notes and phonemes, as well as adapt to different genres.
Moreover, although we have endeavored to incorporate cultural understandings of poetry in different languages, we acknowledge the need for deeper cultural considerations. For example, we noticed that cultural similarities might have an impact on the extent of semantic similarities. This is demonstrated in an English translation of ``MIC Drop'', a K-pop song by BTS originally written in Korean, made by YouTuber Iris Phuong. 
The translated singable lyrics do not include a translation of the term ``hyodo (효도, taking care of parents)'' as there is no direct equivalent in English, while the Japanese version of the song effortlessly conveys the concept as ``koukou (孝行)''. In the future, we aim to expand our dataset to contribute more to lyric translation studies and to further explore the impact of genre and cultural influences on lyric translation.
\end{CJK}

\section{Acknowledgments}

This work was supported in part by JST CREST Grant Number JPMJCR20D4.

\bibliography{ISMIRtemplate}

%
%
%
%
%

\end{document}